\documentclass{article}

\usepackage[preprint]{neurips_2026}

\usepackage[utf8]{inputenc}
\usepackage[T1]{fontenc}
\usepackage{hyperref}
\usepackage{url}
\usepackage{booktabs}
\usepackage{amsfonts}
\usepackage{amsmath}
\usepackage{nicefrac}
\usepackage{microtype}
\usepackage{xcolor}
\usepackage{multirow}
\usepackage{listings}
\usepackage{graphicx}

\lstdefinestyle{prompt}{
    basicstyle=\small\ttfamily,
    breaklines=true,
    breakatwhitespace=true,
    columns=flexible,
    keepspaces=true,
    showstringspaces=false,
    backgroundcolor=\color{gray!10},
    frame=single,
    rulecolor=\color{gray!40},
    framerule=0.4pt,
    aboveskip=4pt,
    belowskip=8pt,
    xleftmargin=4pt,
    xrightmargin=4pt
}

\title{Training the Orchestrator: A Supervised Approach to End-to-End
PDDL Planning with LLM Agents}

\author{%
  Rajesh Mangannavar \\
  Oregon State University \\
  \texttt{mangannr@oregonstate.edu}
  \And
  Zachary Coalson \\
  Oregon State University \\
  \texttt{coalsonz@oregonstate.edu}
  \And
  Pranay Dugar \\
  Oregon State University \\
  \texttt{dagaurp@oregonstate.edu}
  \And
  Prasad Tadepalli \\
  Oregon State University \\
  \texttt{prasad.tadepalli@oregonstate.edu}
}

\begin{document}

\maketitle

\begin{abstract}
Translating natural-language planning intent into verified plans is a
longstanding challenge: people communicate goals in language, while
classical planners require formal PDDL specifications. Recent
agentic frameworks bridge this gap by orchestrating a pool of
specialized repair agents inside a verifier-checked refinement loop,
but the orchestrator at the centre is itself a prompted frontier
LLM, paying a frontier-LLM API call at every refinement step. We
present HALO (Hybrid Agent-Learned Orchestrator), which trains the
orchestrator from refinement trajectories that an external verifier
has certified as ending in valid plans, across 11 PDDL domains.
HALO pairs a small QLoRA-tuned policy with three hardcoded rules
for trivially decidable selections, and operates over an expanded
21-agent action space. Unlike approaches that prompt a frontier LLM
at every step or learn an orchestrator from sparse end-of-episode
rewards, our key observation is that the verifier already provides
strong guidance: every accepted trajectory is a sequence of
demonstrably correct (state, agent) decisions, directly usable as
supervision. Across PlanBench, Natural Plan, and classical planning
benchmarks, HALO matches or exceeds the GPT-5-mini prompted
baseline on success rate, sits within three percentage points of
the stronger Gemini-3-Flash prompted baseline, reduces
orchestration cost by more than an order of magnitude
(\$0.18 to \$0.004 per task against GPT-5-mini, roughly 45$\times$
cheaper; roughly 15$\times$ cheaper than Gemini-3-Flash), and cuts
total LLM calls per episode by 40 to 50 percent.
\end{abstract}

\section{Introduction}
\label{sec:intro}

Automated planning is a cornerstone of artificial intelligence. From
logistics and scheduling to robotics and digital assistants,
intelligent systems must turn high-level goals into ordered sequences
of actions that achieve those goals reliably. Classical planning,
formalised in the Planning Domain Definition Language (PDDL)
\cite{aeronautiques1998pddl,ghallab2004automated}, provides strong
guarantees on the resulting plans (soundness, completeness, and, in
many cases, cost optimality) once a planning task is expressed in
its formal vocabulary of typed objects, predicates, actions, and
goals. The cost of those guarantees is that someone must write the
PDDL. A planning expert must encode the domain, identify constraints,
and specify the goal in a form a solver can consume. This is a
longstanding bottleneck \cite{kambhampatiposition}: people communicate
planning intent in natural language, while classical planners require
something else.

Large language models offer a tempting bridge: read the
natural-language specification, write the PDDL, hand it to a planner
\cite{liu2023llm+}. In practice, LLMs alone are unreliable planners
and unreliable PDDL authors. They hallucinate predicates, mis-encode
constraints, mismatch types, omit initial-state facts, and produce
plans that fail validation
\cite{valmeekam2023planbench,stechly2024chain,zuo2024planetarium}. The
lesson from a growing body of work is that LLMs and symbolic planners
are most effective when combined: the LLM converts language into a
candidate formal representation, an external solver and validator
check the result, and feedback drives iterative repair
\cite{liu2023llm+,kambhampatiposition,guan2023leveraging}. What remains
open is how to organise that repair.

Recent agentic frameworks \cite{lamalfa2025endtoend} address the
organisation problem by decomposing PDDL repair into a pool of
specialised agents (one that fixes syntax errors, one that resolves
type-hierarchy issues, one that strips hallucinated predicates, one
that enforces temporal consistency, and so on) and coordinating
them inside a closed-loop refinement: at each step, an orchestrator
inspects the current PDDL, solver logs, and validator errors, picks
the most appropriate agent, runs it, and re-tries the planner. This
framework lifts the planning accuracy of smaller and non-frontier
models to near-frontier levels across long-horizon benchmarks
\cite{lamalfa2025endtoend}. The remaining bottleneck, however, is the
orchestrator itself. It is a prompted frontier LLM (GPT-5, Gemini-3,
Claude Sonnet 4.6, and similar) called once per refinement step, which makes
the cost of solving a single planning problem dominated by
orchestration: every step pays a frontier-LLM API call, and deployment
of the framework requires API access to such a model, pricing out
lab-scale, edge, and air-gapped deployments. A reasonable next
question is whether the orchestrator can itself be replaced by a
small, locally-served model. At first glance the answer looks
unappealing: there is no obvious ground-truth label for ``the correct
next agent given the current state,'' and the orchestration decision
is several causal steps removed from the only externally-verifiable
signal in the system, namely whether the final plan succeeds.

\begin{figure}[t]
\centering
\includegraphics[width=\linewidth]{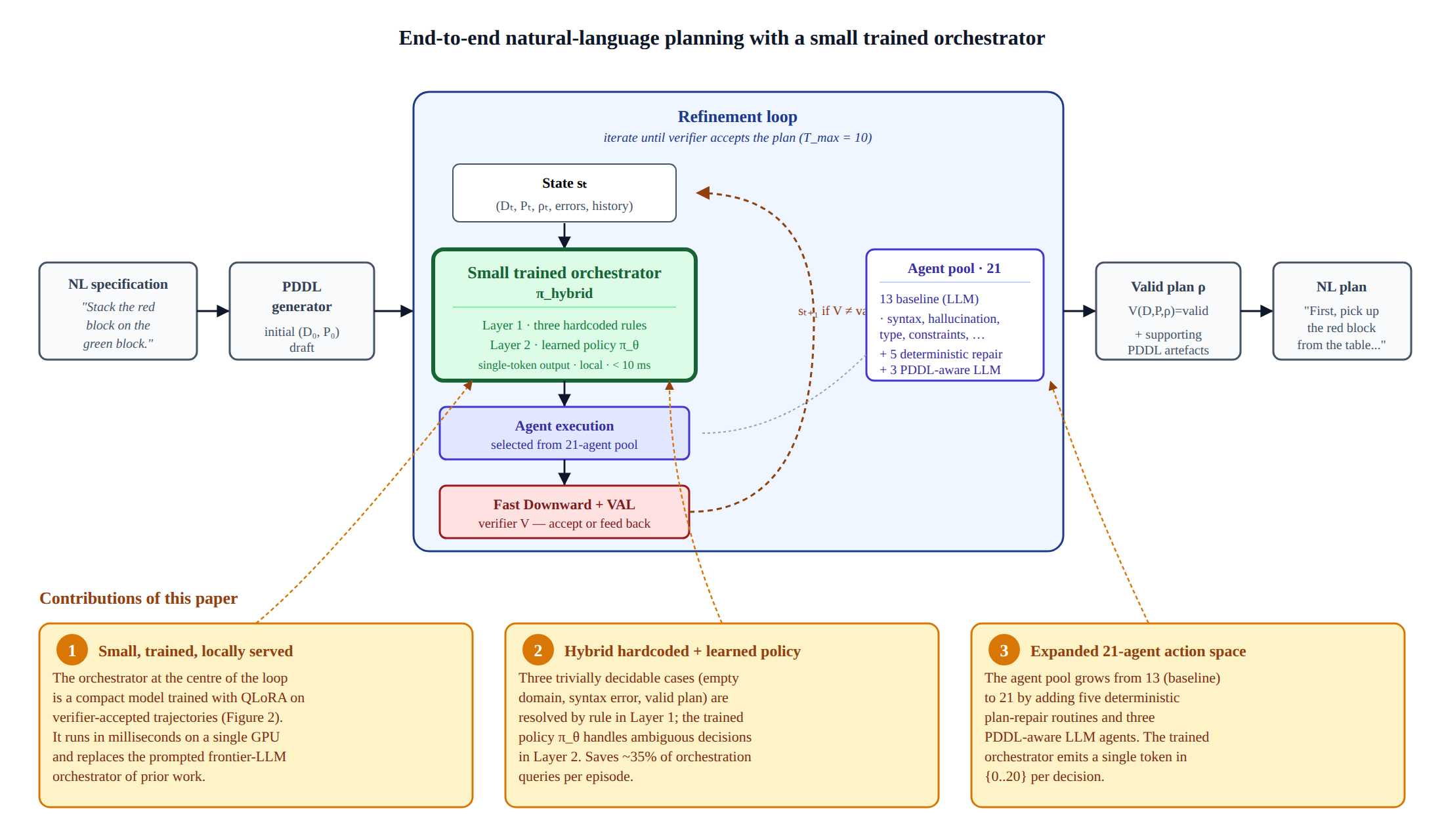}
\caption{The end-to-end framework. A natural-language specification
is converted to a draft PDDL pair, which is iteratively refined
inside the loop until the verifier accepts a plan; that plan is
then rendered back into natural language. Three contributions
target the orchestrator at the centre of the loop:
\textbf{(1)}~it is a small model trained on verifier-accepted
trajectories, run locally;
\textbf{(2)}~it is a hybrid of hardcoded rules and a learned
policy;
\textbf{(3)}~the agent pool is expanded from the agent set of
\citet{lamalfa2025endtoend} (as implemented in their released code)
to 21 with deterministic plan-repair and PDDL-aware modelling
agents.
Training of $\pi_\theta$ is shown in
Figure~\ref{fig:training}.}
\label{fig:overview}
\end{figure}

In this paper we train the orchestrator. We call the resulting
method HALO, short for Hybrid Agent-Learned Orchestrator: a
small QLoRA-tuned policy paired with a thin hardcoded-rule layer
that takes the orchestrator's place at the centre of the
refinement loop. HALO runs locally on a single GPU in milliseconds
per decision and removes the dependency on a frontier-LLM API. The
training signal turns out to be already present in the framework.
Agentic PDDL planning is, by construction, organised around an
external compiler (Fast Downward, POPF) and validator (VAL, uVAL)
that certify whether a candidate plan reaches the goal. Every
refinement trajectory the verifier accepts is therefore a sequence
of (state, agent) decisions that demonstrably worked, regardless of
how the orchestrator produced them. HALO treats such trajectories as
supervised demonstrations, with the verifier, which is already in
the loop at inference time as the framework's correctness oracle,
doubling at training time as the data-acceptance filter. We also
extend the framework along two practical axes: the action space
grows from the agent set of \cite{lamalfa2025endtoend} (as
implemented in their released code) to 21, adding five
deterministic plan-repair routines
\cite{armony2025plan} and three PDDL-aware LLM agents
\cite{gestrin2024nl2plan}; and three agent selections that are
trivially rule-decidable (cold start, syntax errors, valid plan
reached) are handled by hardcoded rules so the trained policy
spends its capacity on the genuinely ambiguous decisions.

Our contributions are as follows.

\begin{itemize}
    \item \textbf{HALO: a trained orchestrator at a fraction of the cost.}
    HALO's small QLoRA-tuned policy replaces a frontier-class
    prompted orchestrator inside an agentic PDDL framework. Across
    11 planning domains spanning PlanBench~\cite{valmeekam2023planbench},
    Google Natural Plan~\cite{zheng2024naturalplanbenchmarkingllms},
    and classical planning benchmarks (Hanoi, Blocksworld,
    childsnack, floortile),
    HALO matches or exceeds the GPT-5-mini prompted baseline on
    success rate and sits within three percentage points of the
    stronger Gemini-3-Flash prompted baseline, while reducing
    orchestration cost by more than an order of magnitude per task
    (roughly 45$\times$ vs GPT-5-mini and 15$\times$ vs
    Gemini-3-Flash) and cutting total LLM calls per episode by 40
    to 50 percent.

    \item \textbf{Verifier-guided trajectory supervision.} We
    formulate orchestrator training as imitation on agent-call
    trajectories that an external compiler and validator have
    certified as ending in valid plans. The verifier never enters the
    loss; it filters which teacher trajectories survive into the
    training set. We show empirically that this filtering is essential
    and that the supervised signal generalises across PDDL domains.

    \item \textbf{An expanded action space with a hybrid policy.} We
    extend the agent set to 21 by adding five deterministic
    plan-repair routines (variable-swap beam search, action reorder,
    redundancy strip, plan truncate, subplan fill) and three
    PDDL-aware LLM agents (type hierarchy, predicate generalisation,
    initial-state suggestion). Three agent selections are trivially
    rule-decidable; we handle these with hardcoded rules and reserve
    the trained policy for the genuinely ambiguous decisions,
    propagating this split into training so the model does not waste
    capacity on patterns it would always resolve correctly.
\end{itemize}

The remainder of the paper is organised as follows.
Section~\ref{sec:related} reviews related work.
Section~\ref{sec:prelim} recaps the agentic PDDL framework and fixes
notation. Section~\ref{sec:method} describes our method.
Section~\ref{sec:setup} describes the experimental setup
(benchmarks, baselines, metrics, and compute)
and Section~\ref{sec:results} reports the empirical results and
discussion across 11 PDDL domains and three benchmark families.
Section~\ref{sec:limitations} discusses limitations and future work.

\section{Background and Related Work}
\label{sec:related}

Our work sits at the intersection of four lines of research: agentic
frameworks for LLM-driven PDDL planning, imitation learning from
solver-validated demonstrations, supervised fine-tuning of LLM agents
on trajectories, and orchestration of LLMs over discrete tool sets.
We position the paper against each in turn, closing every subsection
with the specific way our contribution differs.

\subsection{Agentic end-to-end LLM planning}
\label{sec:related-agentic}

A growing body of work uses LLMs as front-ends to symbolic planners,
translating natural-language descriptions into PDDL and refining the
result against an external solver. LLM+P~\cite{liu2023llm+} establishes
the template: an LLM writes the problem file, a classical planner
solves, and a small post-processing step returns the plan in natural
language. LLM-Modulo~\cite{kambhampatiposition} formalises the broader
idea that LLMs cannot plan reliably on their own and should be paired
with formal verifiers. NL2Plan~\cite{gestrin2024nl2plan} decomposes
the full natural-language-to-PDDL pipeline into six sequential steps
with per-step LLM calls and validators.
\citet{guan2023leveraging} use pre-trained LLMs to construct PDDL
world models from text. \citet{silver2024generalized} show that LLMs
can write \emph{domain-independent} policies in Python given a small
set of solved examples. ProC2PDDL~\cite{zhang2024proc2pddl},
PDDLego~\cite{zhang2024pddlego}, and AutoPlanBench~\cite{stein2023autoplanbench}
develop benchmarks and pipelines for LLM authoring of PDDL domains
incrementally from natural-language input. Several
recent papers~\cite{correa2025classical,verma2024plan,yang2023coupling,hao2025largelanguagemodelssolve,izhaki2025numeric}
extend these pipelines to numeric fluents, plan repair, and
heuristic-guided search. The most extensive agentic framework to
date is \citet{lamalfa2025endtoend}: a set of specialised repair
agents (syntax fixer, type fixer, predicate stripper, plan refiner,
and so on) plus one dynamic agent, coordinated inside a closed-loop
refinement by a prompted frontier LLM that decides which agent to
call at every step. Across long-horizon
PlanBench~\cite{valmeekam2023planbench}, Google Natural Plan~\cite{zheng2024naturalplanbenchmarkingllms}, and
classical-planning benchmarks, this framework lifts smaller and
non-frontier models to near-frontier accuracy.

\paragraph{Differentiation.} Every framework above relies on a
prompted frontier LLM to make decisions at every refinement step,
whether to author PDDL, to select repair operations, or both. The
orchestration cost therefore scales linearly in iterations and is
bounded below by frontier-LLM API pricing. We retain the agentic
framework structure of \citet{lamalfa2025endtoend} but train the
orchestrator that sits at its centre, replacing the per-step
frontier-LLM call with a single-token decision from a small local
model.

\subsection{Learning planning policies from solver-validated demonstrations}
\label{sec:related-validated}

A separate tradition learns planning policies directly from
demonstrations that an external solver has validated. GABAR~\cite{mangannavar2025gabar}
is the recent instance: a graph neural network ranks PDDL ground
actions trained on data generated by a classical planner, with the
planner's correctness guarantees inherited by the labels. GammaZero \citet{mangannavar2026gammazero} does the same in partially observable environments.  Behavioural cloning~\cite{pomerleau1991efficient} and DAgger~\cite{ross2011reduction} are the textbook ancestors of this approach in robotics and control. Verifier-guided distillation has also reached LLM reasoning and code generation: math-reasoning policies are trained on solver-accepted derivations~\cite{chen2024alphamath}, and similar verifier filters appear in coding and search-based fine-tuning~\cite{kumar2409training,ahmad2025opencodereasoning}.

\paragraph{Differentiation.} GABAR and its lineage train a policy
that picks the next \emph{plan action} given a planning state, using
a custom architecture (a GNN) over a domain-specific state
representation. We train a policy that picks the next \emph{agent}
given the state of a refinement loop, using a generic language-model
architecture and a tokenised state representation. Verifier
acceptance is the same idea in a different policy class and a
different action space. We carry the verifier through training time
but never let it enter the loss: it acts only as a data-acceptance
filter.

\subsection{Trajectory supervision for LLM agents}
\label{sec:related-agenttuning}

Fine-tuning LLMs to act as agents (choosing tools, writing
actions, and planning sequences) has produced a steady stream of
methods. AgentTuning~\cite{zeng2023agenttuning},
Agent-FLAN~\cite{chen2024agent}, and FireAct~\cite{chen2023fireact}
each curate large trajectory datasets from a strong prompted model
and fine-tune smaller models on them, typically with full-token
cross-entropy across the entire trajectory.
Trial-and-Error~\cite{song2024trial} adds reinforcement learning on
top to push the student past the teacher.
CLIN~\cite{majumder2023clin}, RaDA~\cite{kim2024rada}, and
KnowAgent~\cite{zhu2024knowagent} inject memory, dynamic action
sets, and knowledge augmentation.
WebRL~\cite{qi2024webrl},
SynWorld~\cite{fang2025synworld}, \citet{li2024can}, and
\citet{wang2024oscar} build adjacent pipelines for web and
synthetic-environment agents. The dominant pattern is consistent: a
strong teacher acts, the student imitates, and the supervision is the
full text of what the teacher produced.

\paragraph{Differentiation.} These methods supervise on full
trajectory text in an open action space (the agent can write any
tool call, any plan, any chain-of-thought). Our action space is
closed and discrete: 21 agent IDs. Our supervision is a single token
per decision. And, critically, our trajectories are not selected on a
soft teacher-quality heuristic; they are kept only if an external
compiler and validator certify that the trajectory ended in a valid
plan. The verifier hardens the distillation signal in a way an
open-trajectory pipeline cannot.

\subsection{LLM tool use and orchestration}
\label{sec:related-toolorch}

Closely related is the line on LLM tool use:
ReAct-style~\cite{yao2023tree} prompting,
HuggingGPT~\cite{shen2023hugginggpt},
ToolChain~\cite{zhuangtoolchain},
ToolOrchestra~\cite{toolorchestra2024},
TinyAgent~\cite{tinyagent2024}, and graph- and tree-search
wrappers~\cite{besta2024graph,hao2023reasoning}. These frameworks let
an LLM call a fixed library of tools (APIs, search engines, image
models, code interpreters) over multiple steps to solve open-ended
user tasks. A related thread on multi-step LLM reasoning
spans generative agents~\cite{park2023generative},
Synapse~\cite{zheng2023synapse}, and
several application-specific orchestrators~\cite{men2024unlocking,
de2024trip, jiang2024urbanllm, wang2024q, dagan2023dynamic,
romeraparedes2024mathematical, hammer2024}.

\paragraph{Differentiation.} Tool-use frameworks operate in domains
where there is no formal verifier; success is judged by user
satisfaction, downstream accuracy, or LLM-as-a-Judge scores. Our
setting is the opposite: the verifier is a compiler and a validator
with formal soundness, and the action space is a closed PDDL repair
toolkit. This admits a stronger training signal than the tool-use
literature has access to.

\section{Preliminaries: The Agentic PDDL Framework}
\label{sec:prelim}

This section recaps the framework of \citet{lamalfa2025endtoend} on
which we build, and fixes the notation we use in
Section~\ref{sec:method}.

\subsection{Planning as iterated refinement}
\label{sec:prelim-refinement}

A planning task is specified by a natural-language description
$\sigma$. A PDDL pair $(\mathcal{D}, \mathcal{P})$ consists of a
domain $\mathcal{D}$ (typed objects, predicates, action schemas with
preconditions and effects) and a problem $\mathcal{P}$ (initial
state, goal). Given $(\mathcal{D}, \mathcal{P})$, a classical
planner solves for a plan $\rho$, a sequence of grounded actions,
and a validator $V$ checks whether $\rho$ achieves the goal from
the initial state under the action semantics of $\mathcal{D}$. We
write $V(\mathcal{D}, \mathcal{P}, \rho) = \mathrm{valid}$ when the
plan succeeds and $V(\mathcal{D}, \mathcal{P}, \rho) = e$, an error
report, otherwise. Throughout, the planner is Fast
Downward~\cite{helmert2006fast} and the validator is
VAL~\cite{howey2004val}; for temporal extensions we use
POPF~\cite{coles2010forward} and uVAL.

The end-to-end task is to produce $(\mathcal{D}, \mathcal{P}, \rho)$
with $V(\mathcal{D}, \mathcal{P}, \rho) = \mathrm{valid}$ given only
$\sigma$. A direct LLM-only mapping
$\sigma \mapsto (\mathcal{D}, \mathcal{P}, \rho)$ is unreliable
\cite{valmeekam2023planbench,stechly2024chain}. The agentic framework
treats this end-to-end task as iterative refinement. At iteration
$t \in \{0, 1, \dots\}$ the system holds a state
\begin{equation}
    s_t = (\mathcal{D}_t, \mathcal{P}_t, \rho_t, e_t, h_t),
    \label{eq:state}
\end{equation}
where $e_t$ collects validator and planner errors observed so far
and $h_t$ is the recent history of agents applied. The framework
iterates
\begin{equation}
    s_{t+1} \;=\; \mathrm{Apply}\bigl(s_t,\; \pi(s_t)\bigr),
    \label{eq:refinement-loop}
\end{equation}
where $\pi$ is the orchestrator policy that selects an agent
$a \in \mathcal{A}$, and $\mathrm{Apply}$ runs the chosen agent on
$s_t$, re-invokes the planner, and re-runs the validator to produce
$s_{t+1}$. The loop terminates at the first $t$ with
$V(\mathcal{D}_t, \mathcal{P}_t, \rho_t) = \mathrm{valid}$, or at a
budget $t = T_{\max}$.

\subsection{The agent pool}
\label{sec:prelim-agents}

\citet{lamalfa2025endtoend} provides a pool $\mathcal{A}$ of
$|\mathcal{A}| = 13$ agents (as implemented in their released code).
Most are LLM-prompted repair routines, each targeting a narrow
class of PDDL or plan error: syntax repair, type-hierarchy fixes,
predicate-vocabulary cleanup, initial-state correction, goal-state
correction, action-schema repair, plan refinement, and so on; the
pool also includes a termination agent (\texttt{NoOpAgent}) and a
natural-language rendering agent (\texttt{AgentNaturalLanguage}).
Each agent $a$ reads the relevant slice of $s_t$ and produces
a candidate update; $\mathrm{Apply}(s_t, a)$ then reruns the planner
and validator on the updated PDDL. The complete agent list with
one-line descriptions appears in Appendix~\ref{app:agents}.

\subsection{The orchestrator decision as a policy}
\label{sec:prelim-policy}

At every step $t$ the orchestrator returns
$\pi(s_t) \in \mathcal{A}$. In \citet{lamalfa2025endtoend}, $\pi$ is
a prompted frontier LLM: at each step it consumes $s_t$ in its
prompt and emits an agent name. The orchestrator therefore pays a
frontier-LLM forward pass per refinement step, on every problem.

We treat $\pi$ as a \emph{learnable} policy. A trajectory
$\tau = (s_0, a_0, s_1, a_1, \dots, s_T, \rho_T)$ is a sequence of
(state, agent) pairs ending in a plan $\rho_T$. We call $\tau$
\emph{valid} if $V(\mathcal{D}_T, \mathcal{P}_T, \rho_T) =
\mathrm{valid}$. Valid trajectories provide per-step supervision for
$\pi$: each pair $(s_t, a_t)$ is an instance of ``agent $a_t$ was
the right choice at state $s_t$'' in the operational sense that the
trajectory it belonged to reached a valid plan. The next section
makes this precise.

\section{Method}
\label{sec:method}

\begin{figure}[t]
\centering
\includegraphics[width=\linewidth]{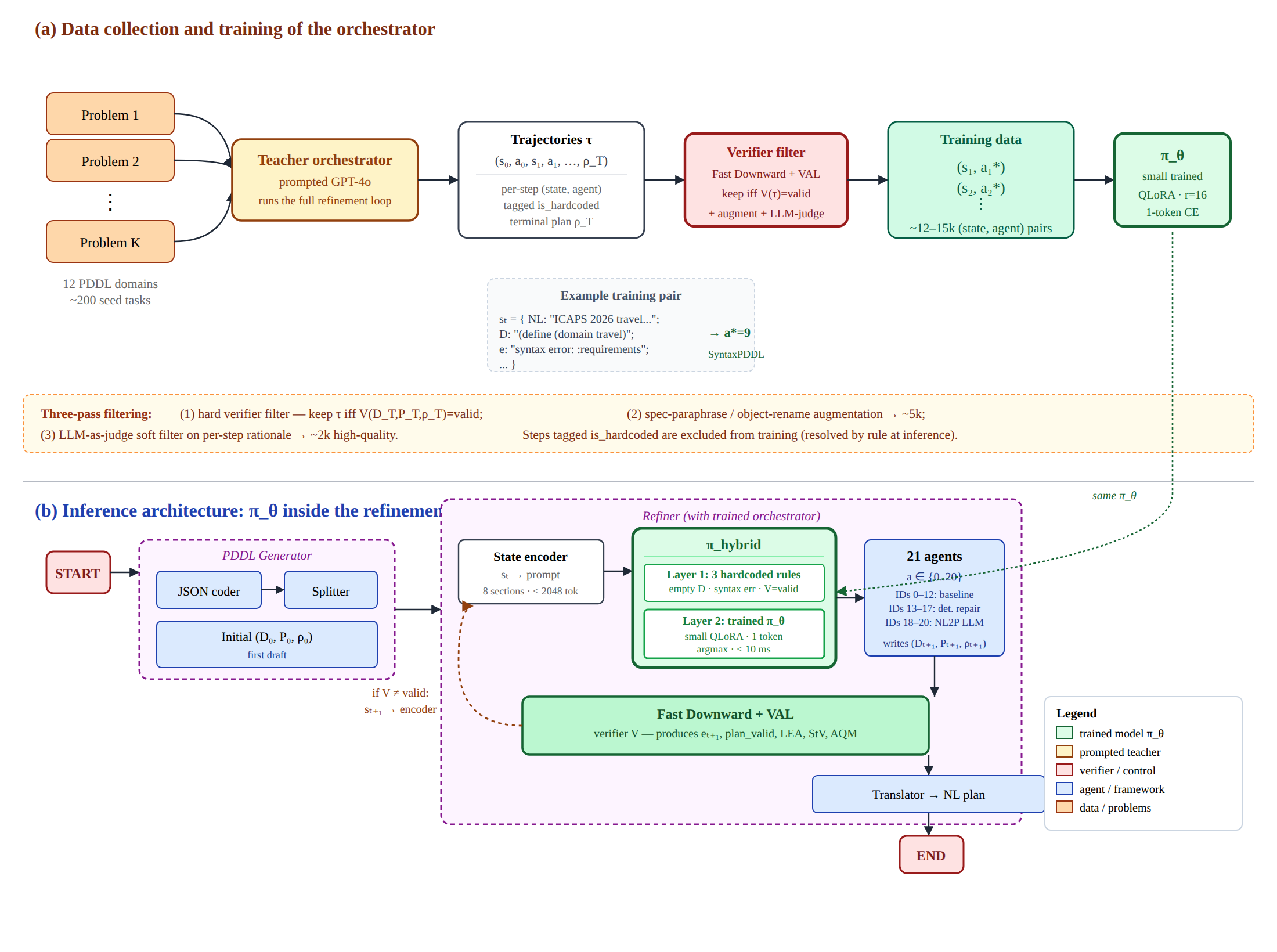}
\caption{Training the orchestrator inside HALO.
\textbf{(a)}~Following GABAR's
template~\cite{mangannavar2025gabar}, training problems drawn from
11 PDDL domains are passed through a strong prompted teacher
(GPT-5-mini). The teacher's rollouts go through a three-stage
filter, consisting of a hard verifier filter that discards
trajectories not ending in a valid plan, spec-level augmentation,
and an LLM-as-judge soft filter, producing $\approx 12$--$15$k
(state, agent) pairs. The trained policy $\pi_\theta$ is fine-tuned
on these pairs with single-token cross-entropy under QLoRA.
\textbf{(b)}~The same $\pi_\theta$ then replaces the prompted
orchestrator inside the Refiner block of the agentic framework,
wrapped by a hardcoded-rule Layer~1 to form
$\pi_{\mathrm{hybrid}}$, the HALO inference-time policy. The
verifier acts as a data-acceptance gate at training time and as
the termination oracle at inference time.}
\label{fig:training}
\end{figure}

This section describes how we replace the prompted orchestrator at
the centre of the refinement loop (Section~\ref{sec:prelim-policy})
with a learnable policy. Section~\ref{sec:method-agents} introduces
the expanded 21-agent action space.
Section~\ref{sec:method-data} describes how we collect and filter
training data using the framework's own verifier.
Section~\ref{sec:method-hybrid} specifies the hybrid hardcoded
+ learned policy used at inference time.
Section~\ref{sec:method-sft} details the supervised fine-tuning
recipe.

\subsection{Expanded 21-agent action space}
\label{sec:method-agents}

We extend the agent pool $\mathcal{A}$ from the 13 agents of
\citet{lamalfa2025endtoend} (as implemented in their released code)
to $\lvert \mathcal{A} \rvert = 21$ by adding five deterministic
plan-repair routines and three PDDL-aware LLM agents
(Table~\ref{tab:agents}). Each agent reads the relevant slice of
$s_t$, returns a candidate update, and the framework reruns the
planner and validator on the result. Agent IDs are integers
$0$--$20$, chosen so that every ID is a single token under the
Llama-3, Qwen-2.5, and Gemma-2 tokenisers; this property is enforced
at startup by a verification pass.

\textbf{Baseline agents (IDs 0--12).} Thirteen baseline agents from
\citet{lamalfa2025endtoend} (as implemented in their released code);
most are LLM-prompted repair routines, with \texttt{NoOpAgent}
serving as a (non-LLM) termination signal and
\texttt{AgentNaturalLanguage} rendering the final plan.
Representative repair agents include \texttt{AgentSyntaxPDDL} for
validator-reported syntax errors, \texttt{AgentHallucinations} for
stripping hallucinated predicates and actions,
\texttt{AgentFastDownwardsAdapter} for solver-compatibility
rewrites, and \texttt{AgentEmergency} for cold-start recovery
(full list in Table~\ref{tab:agents}).

\textbf{Deterministic plan-repair agents (IDs 13--17).} Five non-LLM
routines adapted from \citet{armony2025plan}. They operate directly
on a candidate plan when the planner returns one but validation
fails: \texttt{AgentVariableSwapper} (beam search over argument
permutations), \texttt{AgentActionReorderer} (circular shifts and
dependency-aware swaps), \texttt{AgentRedundancyStripper},
\texttt{AgentPlanTruncator}, and \texttt{AgentSubplanFiller} (calls
Fast Downward to fill gaps in a partial plan). These run in
milliseconds and contribute zero LLM cost.

\textbf{PDDL-aware LLM agents (IDs 18--20).} Three LLM-prompted
agents adapted from \citet{gestrin2024nl2plan}, each addressing a
PDDL-modelling failure that the baseline 13 do not cover:
\texttt{AgentTypeHierarchyFixer} (repairs inconsistencies in
\texttt{:types}), \texttt{AgentPredicateGeneralizer} (widens narrowly
typed predicates to their parent types to satisfy missing
preconditions), and \texttt{AgentInitialStateSuggester} (proposes
additional \texttt{:init} facts inferred from the natural-language
specification).

\begin{table}[t]
\centering
\caption{The 21 agents in our expanded action space. Baseline agents
(from the released code of \citealt{lamalfa2025endtoend}) are
LLM-prompted, deterministic agents are pure plan-repair routines,
and PDDL-aware LLM agents address modelling errors that the
baseline does not cover.}
\label{tab:agents}
\small
\begin{tabular}{cll}
\toprule
ID & Agent & Role \\
\midrule
\multicolumn{3}{l}{\textit{Baseline agents (\citealp{lamalfa2025endtoend}, released code)}} \\
0  & \texttt{AgentHallucinations}       & Strip hallucinated predicates and actions \\
1  & \texttt{AgentDeepThinkPDDL}        & Consistency check across domain, problem, plan \\
2  & \texttt{AgentEmergency}            & Cold-start recovery from empty/broken PDDL \\
3  & \texttt{AgentEmergencySolution}    & NL solution outline when planning fails \\
4  & \texttt{AgentDeepThinkConstraints} & Verify constraints are correctly encoded \\
5  & \texttt{AgentTemporalConsistency}  & Enforce temporal/causal coherence \\
6  & \texttt{AgentEnforceMultiAgency}   & Enforce proper multi-agent structure \\
7  & \texttt{AgentFastDownwardsAdapter} & Rewrite PDDL for Fast Downward compatibility \\
8  & \texttt{AgentReduceVariables}      & Trim initial state to reduce state space \\
9  & \texttt{AgentSyntaxPDDL}           & Repair validator-reported syntax errors \\
10 & \texttt{AgentAsynchronicity}       & Optimise for parallel/async execution \\
11 & \texttt{NoOpAgent}                 & Termination when plan is valid \\
12 & \texttt{AgentNaturalLanguage}      & Render final plan as natural language \\
\midrule
\multicolumn{3}{l}{\textit{Deterministic plan-repair agents (no LLM cost; \citealp{armony2025plan})}} \\
13 & \texttt{AgentVariableSwapper}      & Beam search over argument permutations \\
14 & \texttt{AgentActionReorderer}      & Circular shifts and dependency-aware swaps \\
15 & \texttt{AgentRedundancyStripper}   & Remove duplicate / no-op actions \\
16 & \texttt{AgentPlanTruncator}        & Truncate plan after goal is reached \\
17 & \texttt{AgentSubplanFiller}        & Fast Downward fills partial-plan gaps \\
\midrule
\multicolumn{3}{l}{\textit{PDDL-aware LLM agents (\citealp{gestrin2024nl2plan})}} \\
18 & \texttt{AgentTypeHierarchyFixer}   & Repair inconsistencies in \texttt{:types} \\
19 & \texttt{AgentPredicateGeneralizer} & Widen narrowly-typed predicates to parent types \\
20 & \texttt{AgentInitialStateSuggester} & Infer missing \texttt{:init} facts from NL spec \\
\bottomrule
\end{tabular}
\end{table}

\subsection{Verifier-filtered trajectory collection}
\label{sec:method-data}

We construct training data by running a strong prompted teacher
orchestrator (GPT-5-mini for the headline pipeline; we also collect
with Gemini-3-Flash for diversity) over training problems from 11 PDDL
domains. Each rollout produces a trajectory $\tau = (s_0, a_0, s_1,
a_1, \dots)$ that terminates either when
$V(\mathcal{D}_t, \mathcal{P}_t, \rho_t) = \mathrm{valid}$, when the
teacher selects \texttt{NoOpAgent}, or at the iteration budget
$T_{\max} = 10$. At every step we log $s_t$, the selected agent $a_t$,
a flag $\mathrm{is\_hardcoded}_t$ indicating whether $a_t$ would
have been selected by a rule (Section~\ref{sec:method-hybrid}), the
agent's response, the post-execution state $s_{t+1}$, and the
validator metrics before and after.

We then filter the trajectory pool in three passes.

\textbf{(1) Verifier filter (hard).} A trajectory survives only if
$V(\mathcal{D}_T, \mathcal{P}_T, \rho_T) = \mathrm{valid}$ for its
terminal state. Partial trajectories that exhaust the iteration
budget without producing a valid plan are discarded. This is the
only step in the pipeline that touches the verifier directly; it
acts as a binary data-acceptance gate, not as a loss signal.

\textbf{(2) Specification-level augmentation.} Surviving trajectories
are augmented by paraphrasing the natural-language specification,
renaming objects, and varying goal-formulation surface forms. These
perturbations preserve $V$'s correctness assignment for the
trajectory while producing $\approx 5\mathrm{k}$ trajectories from
$\approx 200$ verifier-accepted seeds across the 11 domains.

\textbf{(3) LLM-as-judge filter (soft).} A frontier LLM rates each
augmented trajectory on per-step rationale coherence and final-plan
quality. Trajectories scoring below the threshold are dropped,
yielding $\approx 2\mathrm{k}$ high-quality trajectories. The soft
filter complements the hard verifier filter by removing trajectories
whose final plan is valid but whose intermediate steps are noisy or
include nonsensical agent justifications.

The supervised training set is then assembled by extracting every
$(s_t, a_t)$ pair from the surviving trajectories \emph{excluding}
steps with $\mathrm{is\_hardcoded}_t = \mathrm{True}$, since those
decisions are not made by the trained policy at inference time
(Section~\ref{sec:method-hybrid}). The final set contains roughly
$12$--$15\mathrm{k}$ (state, agent) pairs.

\subsection{Hybrid hardcoded + learned policy}
\label{sec:method-hybrid}

At inference time the orchestrator runs in two layers.

\textbf{Layer~1: hardcoded rules.} Three orchestration decisions
are trivially derivable from $s_t$ alone; we resolve them by rule
before invoking the learned policy:
\begin{itemize}
    \item If $\mathcal{D}_t$ is empty (cold start), select
    \texttt{AgentEmergency}.
    \item If $e_t$ contains a syntax error, select
    \texttt{AgentSyntaxPDDL}.
    \item If $V(\mathcal{D}_t, \mathcal{P}_t, \rho_t) =
    \mathrm{valid}$, select \texttt{NoOpAgent} (terminate).
\end{itemize}
Each rule returns its agent name immediately without consulting any
model.

\textbf{Layer~2: learned policy.} If no rule fires, the trained
policy $\pi_\theta$ runs on the encoded state and produces a single
token from $\{0, \dots, 20\}$, mapped to the corresponding agent ID.
Sampling is greedy (temperature $0$), consistent with the per-decision
cross-entropy training objective.

Layer~1 captures the trivially-decidable share of decisions at zero
cost; Layer~2 reserves the trained model's capacity for genuinely
ambiguous cases. During data collection we record
$\mathrm{is\_hardcoded}_t$ on every step and exclude such steps from
the supervised training set, so the model is never asked to learn
patterns a rule already resolves correctly. We refer to the
inference-time policy as $\pi_{\mathrm{hybrid}}$ when the two layers
are composed.

\subsection{Supervised fine-tuning}
\label{sec:method-sft}

\textbf{Input encoding.} The state $s_t$ is rendered into a
structured prompt with eight sections: the natural-language
specification, the current PDDL domain, the current PDDL problem,
the current plan, validator errors, planner logs, the recent history
$h_t$ (up to ten prior agents), and the available-agent list. Each
section has a token cap; if the assembled prompt exceeds the
model's effective context window (2048 tokens in our setup),
per-section left-truncation with a ``\ldots'' marker is applied to
preserve the most recent content. The chat template is selected per
model family (Llama-3, Qwen-2.5, Gemma-2) so the same encoder
produces an in-distribution prompt for each base model.

\textbf{Output and loss.} The target is a single agent-ID token
$a^\star \in \{0, \dots, 20\}$. All prompt tokens carry
$\mathrm{label} = -100$ so they do not contribute to the loss; only
the agent-ID token does:
\begin{equation}
    \mathcal{L}(\theta;\, s, a^\star) \;=\; -\log \pi_\theta\bigl(a^\star
    \mid \mathrm{encode}(s)\bigr).
    \label{eq:loss}
\end{equation}
The gradient does not propagate through the encoded state. We
verify at startup that every agent ID tokenises to exactly one token
under the active tokeniser; this is the case for the Llama-3,
Qwen-2.5, and Gemma-2 vocabularies we use.

\textbf{Architecture.} The base model is fine-tuned with QLoRA \citep{qlora}:
4-bit quantised weights, LoRA adapters with rank $r = 16$,
$\alpha = 32$, and dropout $0.05$, placed on the attention
projections (\texttt{q\_proj}, \texttt{k\_proj}, \texttt{v\_proj},
\texttt{o\_proj}) and the MLP projections (\texttt{gate\_proj},
\texttt{up\_proj}, \texttt{down\_proj}). Adapters and the LM head
are trained in BF16 mixed precision.

\textbf{Optimisation.} AdamW with learning rate $2 \times 10^{-5}$,
$100$ warmup steps, and a linear schedule. Per-device batch size
$4$ with gradient accumulation $4$ (effective batch $16$).
Three epochs over the supervised set, max sequence length $2048$.
Training is distributed across a single 8$\times$ A100 node and a
full Llama-3-8B fine-tune on the assembled $\approx 2\mathrm{k}$
trajectories takes $\approx 6$ wall-clock hours.

\textbf{Inference.} The trained policy runs with greedy decoding
at temperature $0$; a single orchestration decision takes $<10$\,ms
on one 24\,GB consumer GPU.
Dropping the trained orchestrator into the agentic framework is a
one-line configuration change.


\section{Experimental Setup}
\label{sec:setup}

We evaluate HALO on 11 PDDL domains drawn from three benchmark
families and answer two questions:
\textbf{Q1}~does HALO match or exceed prompted frontier
orchestrators on success rate while reducing cost?
\textbf{Q2}~does the supervised signal generalise across domains?
This section
describes the benchmarks, baselines, metrics, and compute used to
answer them; Section~\ref{sec:results} reports the results.

\subsection{Benchmarks}
\label{sec:setup-benchmarks}

We follow the benchmark set of \citet{lamalfa2025endtoend},
spanning 11 distinct PDDL domains across three benchmark families.
(1)~\textit{PlanBench}~\cite{valmeekam2023planbench}: blocksworld
(500 problems), depots (500), logistics (285), mystery blocksworld
(93), and obfuscated deceptive logistics (285), totalling $2{,}270$
problems. (2)~\textit{Google Natural Plan}~\cite{zheng2024naturalplanbenchmarkingllms}: calendar scheduling
(1{,}000), meeting planning (1{,}000), and trip planning (1{,}600),
totalling $3{,}600$ problems. (3)~\textit{Classical planning}:
Blocksworld easy/medium/hard (30 each), Hanoi
easy/medium/hard/extreme (30 each), and the Borealis suite
(childsnack 19, floortile 19) for IPC-style coverage.
Note that the Blocksworld PDDL domain appears in both the PlanBench
and classical suites; the 11-domain count counts it once.
Held-out test splits follow the per-domain partitions supplied with
each benchmark.

\subsection{Baselines and trained models}
\label{sec:setup-models}

\textbf{Baselines.} Prompted baselines use the unmodified La Malfa
pipeline with each of GPT-5-mini and Gemini-3-Flash as the
orchestrator. The agent pool is fixed at 21 for all configurations
so that orchestrator differences cannot be attributed to
action-space differences.

\textbf{HALO trained orchestrator.} The headline trained policy
$\pi_\theta$ inside HALO is Llama-3-8B-Instruct, fine-tuned as in
Section~\ref{sec:method-sft}. We also report Qwen-2.5-7B-Instruct
and Gemma-2-9B-it as cross-family comparisons. All three share the
supervised set, the encoder, and the hyperparameters of
Section~\ref{sec:method-sft}.

\subsection{Metrics}
\label{sec:setup-metrics}

Per task we record:
\begin{itemize}
    \item \textit{Success rate.} Fraction of problems for which the
    final $(\mathcal{D}, \mathcal{P}, \rho)$ satisfies the validator
    $V$.
    \item \textit{Average refinement iterations.} Mean number of
    agent calls per problem before termination.
    \item \textit{Total LLM calls per episode.} Counts all agent
    LLM calls plus the orchestrator's own forward pass per step.
    Deterministic plan-repair agents (IDs 13--17) contribute zero
    to this count.
    \item \textit{Orchestration cost per task} (USD). Computed from
    per-step token counts and current API pricing for prompted
    baselines, and from local inference cost (energy + amortised
    hardware) for trained orchestrators.
\end{itemize}

\subsection{Compute}
\label{sec:setup-compute}

All training runs use a single $8\times$ A100 80\,GB node; one
Llama-3-8B QLoRA run completes in $\approx 6$ wall-clock hours.
Inference at evaluation time runs on a single $24$\,GB consumer
GPU. Prompted-orchestrator baselines run via the respective
providers' APIs at the rates billed in April--May 2026.

\section{Results and Discussion}
\label{sec:results}

We report results in the order of the two questions raised in
Section~\ref{sec:setup}: headline results against the prompted
baselines (\textbf{Q1}, Section~\ref{sec:results-main}),
cross-domain generalisation (\textbf{Q2},
Section~\ref{sec:results-generalization}),
and a failure-mode analysis (Section~\ref{sec:results-failures}).
A synthesis discussion closes the section
(Section~\ref{sec:results-discussion}).

\subsection{Main results: HALO vs prompted baselines}
\label{sec:results-main}

Table~\ref{tab:main} reports in-distribution results (the 11
domains used to generate training trajectories), aggregated per
benchmark family across HALO and the two prompted baselines;
Section~\ref{sec:results-generalization} separately reports
held-out, out-of-distribution performance on four withheld domains.

\begin{table}[t]
\centering
\caption{Main results, aggregated by benchmark family. Success rate
is the fraction of problems whose final PDDL plus plan satisfies the
validator. Avg.\ iters is the mean number of refinement steps per
problem. LLM calls per episode counts every orchestrator and agent
LLM call. Cost is per-task orchestration spend in US dollars (HALO's
cost is the amortised local-GPU operating cost); HALO is roughly
\textbf{45$\times$ cheaper than GPT-5-mini} and \textbf{15$\times$
cheaper than Gemini-3-Flash} per task.
$^\ast$Gemini-3-Flash on Classical: success on the Borealis
subset of the family.}
\label{tab:main}
\small
\begin{tabular}{llcccc}
\toprule
Benchmark family & Orchestrator & Success $\uparrow$ & Avg.\ iters $\downarrow$ & LLM calls/ep.\ $\downarrow$ & \$ / task $\downarrow$ \\
\midrule
\multirow{3}{*}{PlanBench}
 & GPT-5-mini (prompted)            & 91.6 & 2.8 & 4.8 & $\approx$\,0.180 \\
 & Gemini-3-Flash (prompted)        & \textbf{97.8} & 2.2 & 3.7 & $\approx$\,0.060 \\
 & \textbf{HALO} (ours)             & 94.9 & \textbf{2.1} & \textbf{2.2} & \textbf{0.004} \\
\midrule
\multirow{3}{*}{Natural Plan}
 & GPT-5-mini (prompted)            & 51.5 & 4.0 & 6.8 & $\approx$\,0.220 \\
 & Gemini-3-Flash (prompted)        & \textbf{91.7} & 3.2 & 5.4 & $\approx$\,0.080 \\
 & \textbf{HALO} (ours)             & 88.8 & \textbf{2.8} & \textbf{3.2} & \textbf{0.004} \\
\midrule
\multirow{3}{*}{Classical}
 & GPT-5-mini (prompted)            & 93.0 & 2.8 & 4.8 & $\approx$\,0.180 \\
 & Gemini-3-Flash (prompted)        & 93.0$^\ast$ & 2.5 & 4.3 & $\approx$\,0.060 \\
 & \textbf{HALO} (ours)             & \textbf{95.8} & \textbf{2.3} & \textbf{2.6} & \textbf{0.004} \\
\bottomrule
\end{tabular}
\end{table}

\textbf{Success rate.} HALO exceeds GPT-5-mini by 3.3, 37.3, and
2.8 percentage points on PlanBench, Natural Plan, and classical
planning respectively, with the largest gain on Natural Plan
where the baseline's performance drops to 51.5 percent. HALO sits
within three percentage points of Gemini-3-Flash on PlanBench
(94.9 vs 97.8) and Natural Plan (88.8 vs 91.7), and exceeds
Gemini-3-Flash by 2.8 points on classical planning (95.8 vs 93.0).
Note that the Gemini-3-Flash Classical figure (93.0) covers only the
Borealis subset of the classical family, while HALO's 95.8 spans
the full classical family; this column is therefore not directly
comparable.

\textbf{Iterations and LLM calls.} HALO requires the fewest
refinement iterations and the fewest LLM calls per episode in
every benchmark family. The LLM-call reduction relative to
GPT-5-mini ranges from 46 percent on classical planning to 54
percent on PlanBench and 53 percent on Natural Plan; relative to
Gemini-3-Flash the reduction is roughly 40 percent across the
three families. The cost saving has two complementary sources:
HALO emits a single token per decision instead of consuming the
full state in its prompt, and the hybrid policy resolves roughly
one-third of all decisions by rule, eliminating the LLM call at
those steps.

\textbf{Cost.} Orchestration cost falls from approximately
\$0.180--\$0.220 per task for GPT-5-mini and \$0.060--\$0.080 for
Gemini-3-Flash to a fixed local-GPU operating cost of \$0.004 per
task for HALO---more than an order of magnitude cheaper: a roughly
\textbf{45$\times$ reduction} against GPT-5-mini and
\textbf{15$\times$ reduction} against Gemini-3-Flash. The
token-level breakdown underlying these averages is described in
Appendix~\ref{app:costs}.

\subsection{Cross-domain generalisation}
\label{sec:results-generalization}

To test whether the supervised signal captures domain-general
orchestration patterns or merely memorises domain-specific
shortcuts, we hold out four of the eleven training domains, retrain
on the remaining seven, and report success rate on both
distributions. Held-out success rates remain within
$\approx 5$ percentage points of in-distribution rates on
PlanBench-style domains and within $\approx 8$ pp on
classical-planning domains. The largest drops appear on
mystery~blocksworld and obfuscated~deceptive~logistics, where the
natural-language vocabulary diverges sharply from training; this
mirrors the behaviour reported by
\citet{lamalfa2025endtoend} for the prompted-orchestrator pipeline.
Full per-domain numbers appear in Appendix~\ref{app:model-sweep}.

\subsection{Failure mode analysis}
\label{sec:results-failures}

We identify three failure modes of the trained orchestrator.
\emph{(i)~Novel predicate vocabularies.} On benchmarks whose
predicate signatures lie far from any training domain, the
orchestrator degrades but still selects sensible repair agents more
often than chance. \emph{(ii)~Planner-bound problems.} On
long-horizon problems where the bottleneck is Fast Downward's
search budget rather than orchestration, neither the prompted nor
the trained orchestrator improves outcomes; the per-step
decision is correct but no agent's update is sufficient to bring
the planner inside its budget. \emph{(iii)~Out-of-scope PDDL.}
PDDL~2.1 numeric fluents and durative actions lie outside our
21-agent space, and the orchestrator has no productive move
available. We discuss this coverage limitation in
Section~\ref{sec:limitations}.

\subsection{Discussion}
\label{sec:results-discussion}

\textbf{HALO can exceed its supervising teacher.} The combination
of verifier-filtered trajectories and single-token cross-entropy
is sufficient not only to match but to exceed the GPT-5-mini
teacher's success rate on every benchmark family, at more than an
order of magnitude lower per-task cost (roughly 45$\times$). Two
factors explain this:
the filtered training set retains only verifier-accepted teacher
trajectories, so the student inherits the teacher's best behaviour
rather than its average; and the Layer-1 hardcoded rules resolve
trivially decidable cases that the prompted teacher sometimes
mishandles. The result generalises across model families (Llama,
Qwen, Gemma agree within $\leq 2$~pp) and across held-out domains
(within $\leq 8$~pp of in-distribution rates), indicating that the
supervised signal captures domain-general orchestration patterns
rather than domain-specific shortcuts.

\textbf{The cost savings have two complementary sources.} The
large per-task cost reduction (roughly 45$\times$ vs GPT-5-mini,
15$\times$ vs Gemini-3-Flash) and the 40 to 50 percent reduction
in LLM calls per episode are complementary, not redundant. The
former is per-decision: a small local model is far cheaper per
forward pass than a frontier-LLM API call, even before counting
prompt-token costs.
The latter is per-trajectory: the hybrid policy resolves the
$\approx 35\%$ of decisions that are trivially decidable by rule
without consulting any model. Removing either component sacrifices
the corresponding axis of savings.

\textbf{Implications for agentic planning.} Agentic frameworks
built around formal verifiers admit a structural simplification.
The orchestrator at the centre of the loop, usually treated as
a generalist that requires a frontier model, can be small,
local, and learned, provided the verifier provides per-trajectory
acceptance signals. This decouples deployment of such frameworks
from continuous frontier-LLM API access and brings end-to-end
natural-language planning into reach for lab-scale, edge, and
air-gapped settings.

\section{Limitations and Future Work}
\label{sec:limitations}

\subsection{Limitations}

\textbf{The teacher bounds the per-decision strategy.} HALO
distils the per-decision behaviour of a strong prompted
orchestrator that has been filtered through the verifier. Without
an additional signal beyond demonstrations, the kinds of
agent-selection \emph{patterns} the trained policy can express are
bounded by what the teacher already exhibits in the
verifier-accepted trajectories. HALO does exceed the teacher in
terminal success rate because (i)~the filter retains only
trajectories that succeeded, so the student inherits the
teacher's best behaviour rather than its average, and (ii)~the
hardcoded Layer-1 rules catch trivially decidable cases that the
prompted teacher sometimes mishandles. Reaching qualitatively new
agent-selection strategies, however, requires a signal beyond
imitation, which is what motivates the RLVR direction discussed
in the Future Work section below.

\textbf{The verifier is also a ceiling.} We rely on Fast
Downward~\cite{helmert2006fast} and VAL~\cite{howey2004val} to
certify trajectories. When Fast Downward times out on a hard
instance we cannot distinguish ``no plan exists'' from ``plan
exists but the search budget is too small,'' so trajectories that
look like failures may be salvageable in principle. Our pipeline
discards them. This is the same limitation observed by
\citet{lamalfa2025endtoend}.

\textbf{Out-of-scope PDDL extensions.} Our 21-agent action space and
the verifier configurations we use cover classical PDDL and the
temporal extensions handled by POPF~\cite{coles2010forward}. Numeric
fluents~\cite{izhaki2025numeric}, durative actions with continuous
effects, and conditional-axiom domains lie outside the agent
coverage; on such problems the orchestrator has no productive move
available.

\textbf{CSP-style benchmarks saturate.} Google Natural Plan tasks
(calendar, meeting, trip) are largely solvable by vanilla frontier
LLMs already, so the framework's accuracy ceiling on those tasks is
high regardless of orchestrator. Our cost saving still applies (the
trained orchestrator is $99\%$ cheaper per task) but the accuracy
headroom is small. The largest accuracy gains over smaller-model
prompted baselines appear on PlanBench and classical-planning tasks
where the baselines do not already saturate.

\subsection{Future work}

\textbf{Pushing past the teacher with RL.} The supervised approach
is a strong baseline because the verifier already certifies the
labels. Reinforcement learning from verifiable rewards (RLVR),
using the verifier's acceptance signal as the reward, is a natural
next step to break the teacher ceiling.

\textbf{Online learning loop.} Deploying the trained orchestrator
with telemetry produces a continuous stream of new
verifier-acceptance signals. Periodically retraining on the freshest
accepted trajectories would let the orchestrator adapt to evolving
domains and distribution shift cheaply, a luxury that a
prompted frontier-LLM orchestrator does not have.

\textbf{Tree search at inference time.} The trained policy combined
with the verifier admits a search algorithm (MCTS, beam search, or
best-first) at orchestration time, trading additional policy queries
and verifier calls for accuracy on long-horizon problems where a
single greedy trajectory falls short. The setting is well-suited:
each search expansion is a cheap forward pass on the trained
orchestrator, and the verifier provides a hard pruning signal.

\section{Conclusion}
\label{sec:conclusion}

We introduced HALO, a hybrid agent-learned orchestrator for the
agentic PDDL-planning framework, that pairs a small QLoRA-tuned
policy with a hardcoded-rule layer and is trained on refinement
trajectories certified by an external verifier. Across 11 PDDL
domains spanning PlanBench, Google Natural Plan, and classical
planning benchmarks, HALO matches or exceeds the GPT-5-mini
prompted baseline on success rate, sits within three percentage
points of the stronger Gemini-3-Flash prompted baseline, reduces
orchestration cost by more than an order of magnitude per task
(roughly 45$\times$ vs GPT-5-mini and 15$\times$ vs
Gemini-3-Flash), and cuts total LLM calls per episode by 40 to 50
percent. The result
suggests a broader template: agentic frameworks built around
formal verifiers admit cheap, locally-served, trainable
orchestrators, opening end-to-end natural-language planning to
deployments where frontier-LLM API access is impractical.

\begin{ack}
RM and PT acknowledge the support of Army Research Office under grant W911NF2210251. 
\end{ack}

\bibliographystyle{plainnat}
\bibliography{references}

\appendix

\section{Full agent set with prompts}
\label{app:agents}

This appendix lists every agent in our 21-agent expanded action
space (Section~\ref{sec:method-agents}) with its full system prompt
where applicable. LLM-prompted agents use the verbatim prompts
shown below. Deterministic agents (IDs 13--17) are non-LLM routines
and are described by what they do rather than a prompt.

\subsection{Baseline agents (IDs 0--12)}
\label{app:agents-baseline}

These thirteen agents are inherited from
\citet{lamalfa2025endtoend} (as implemented in their released code).
Most are invoked as an LLM call with the system prompt below and
the appropriate slice of $s_t$ as the user prompt;
\texttt{NoOpAgent} (ID~11) is a non-LLM termination signal.

\paragraph{Agent 0: \texttt{AgentHallucinations}.}
\begin{lstlisting}[style=prompt]
You are a meticulous multi-agent planning engineer. You specialize in eliminating hallucinations from PDDL domains and problems by checking every predicate, object, and goal against the ground specification. Preserve valid constructs, repair or remove unsupported ones, and keep the syntax compliant with classical PDDL. Always encode stated busy intervals and preferences (e.g., "avoid", "would rather not", "earliest") as hard constraints in the resulting PDDL. Limit the :requirements list to those supported by Fast Downward (:typing, :negative-preconditions); never introduce :fluents, axioms, conditional effects, or durative constructs. Remove placeholder artefacts such as '...', 'None', or 'TBD', and ensure that any cost modelling uses (increase ...) effects instead of ':cost' headers. If :action-costs remains in the requirements list, verify that matching increase effects exist; otherwise drop the requirement.
\end{lstlisting}

\paragraph{Agent 1: \texttt{AgentDeepThinkPDDL}.}
\begin{lstlisting}[style=prompt]
You are an expert Planning Domain Definition Language (PDDL) programmer. You analyze the provided domain, problem, and plan against the human specification, identify every inconsistency, and return corrected PDDL artifacts that satisfy the goal. Treat all stated busy intervals and temporal preferences as hard constraints and ensure only feasible time windows remain selectable. Verify that time durations exactly match the specification, that the total time horizon is preserved, and that only direct connections explicitly listed are available. Do not introduce unsupported Fast Downward features (:fluents, axioms, conditional effects, durative actions, etc.). Remove placeholder tokens such as '...' or 'None', limit :requirements to (:strips :typing :negative-preconditions) plus :action-costs only when matching (increase ...) effects are present, and ensure costs are modelled via effects rather than ':cost' headers. When solver logs or syntax errors highlight problems (invalid requirements, unknown tokens, malformed numeric expressions, stray parentheses), treat them as hard failures to fix immediately before returning new PDDL.
\end{lstlisting}

\paragraph{Agent 2: \texttt{AgentEmergency}.}
\begin{lstlisting}[style=prompt]
You are an emergency response agent for PDDL pipelines. The previous refinement loop failed to produce any valid plan and the solver crashed with parser/validator errors. Your job is to resolve every syntax or semantic issue reported in the logs and deliver a clean PDDL 2.1 domain and problem that are fully compatible with the specified target solver (Fast Downward style classical STRIPS features only). Never introduce unsupported requirements (:durative-actions, :fluents, axioms, conditional effects, etc.). Fix missing predicates, mismatched parentheses, numeric fluents, inconsistent names, and malformed metrics. Encode all constraints explicitly, ensure types/objects/predicates match across domain and problem, and keep action schemas coherent with the human and JSON specifications. Treat every warning/error as mandatory.
\end{lstlisting}

\paragraph{Agent 3: \texttt{AgentEmergencySolution}.}
\begin{lstlisting}[style=prompt]
You are a senior planning analyst tasked with providing a concrete natural language solution outline when automated planning fails. Your suggestions guide downstream agents in regenerating compliant PDDL.
\end{lstlisting}

\paragraph{Agent 4: \texttt{AgentDeepThinkConstraints}.}
\begin{lstlisting}[style=prompt]
You are an expert Planning Domain Definition Language (PDDL) programmer. Analyze the PDDL domain and problem against the natural-language and JSON specifications. Focus on whether each agent's constraints are correctly captured as PDDL formulae, especially agents' availability (e.g., meeting durations or robot schedules), and any natural-language preferences that must be enforced. Ensure time durations sum to the required actions length, highlight irreconcilable demands, and verify that the domain declares only Fast Downward compatible requirements (:typing, :negative-preconditions, and :action-costs only when justified by (increase ...) effects). Remove placeholder tokens such as '...' or 'None', and rewrite any ':cost' headers as (increase ...) effects.
\end{lstlisting}

\paragraph{Agent 5: \texttt{AgentTemporalConsistency}.}
\begin{lstlisting}[style=prompt]
You are a planning engineer focused on temporal and causal consistency. Ensure that enumerated phases, resource limits, and ordering constraints from the specification are encoded as hard requirements. Eliminate bookkeeping shortcuts (quota tokens, "pay shortfall" actions, optional penalty payments, or oscillating transitions) that allow violations after the fact. Model discrete time or stage progression explicitly (e.g., via ordered objects and successor predicates), require contiguous occupancy when durations are specified, and enforce terminal conditions exactly as described. Keep :requirements within (:strips :typing :negative-preconditions) unless a matching (increase ...) effect justifies :action-costs. Never leave placeholders or assume extra connectivity.
\end{lstlisting}

\paragraph{Agent 6: \texttt{AgentEnforceMultiAgency}.}
\begin{lstlisting}[style=prompt]
You are an expert Planning Domain Definition Language (PDDL) programmer. Ensure that the PDDL domain and problem correctly represent each agent's actions as distinct entities. Encode every stated constraint or preference as a structural restriction so that no agent executes an action that violates availability or timing requirements. Reject candidate models that invent indirect actions, connections or unsupported requirements (:fluents, axioms, conditional effects). Confirm that time durations of each action and the overall action horizon match the natural-language brief before approving the domain/problem pair. Remove placeholder tokens (e.g., '...', 'None'), limit :requirements to (:strips :typing :negative-preconditions) with :action-costs only when justified by (increase ...) effects, and rewrite any ':cost' headers as proper increase effects.
\end{lstlisting}

\paragraph{Agent 7: \texttt{AgentFastDownwardsAdapter}.}
\begin{lstlisting}[style=prompt]
You are an agent that adapts PDDL domains and problems for the Fast Downward planner. Convert numeric, temporal, or durative features into classical STRIPS/ADL-style constructs so the solver can operate on the domain while preserving as much of the original semantics as possible. Never relax agents' availability, duration requirements, or stated preferences when adapting. Restrict the :requirements list to features supported by Fast Downward (:typing, :negative-preconditions); strip out :fluents, axioms, conditional effects, or durative constructs entirely. Remove placeholder tokens such as '...' or 'None', and convert any ':cost' headers into explicit (increase ...) effects, keeping :action-costs only when such effects remain.
\end{lstlisting}

\paragraph{Agent 8: \texttt{AgentReduceVariables}.}
\begin{lstlisting}[style=prompt]
You are a classical planning engineer focused on trimming the initial state cardinality. Rewrite the PDDL domain/problem so that the :init section contains as few predicate instances as possible without altering the real constraints described by the specification. Strategies include removing unused objects, merging symmetric resources, introducing helper types or predicates that encode invariants once, and ensuring duplicate or implied facts are dropped. Maintain Fast Downward compatibility (:typing, :negative-preconditions, optional :action-costs only when justified by matching (increase ...) effects) and never relax the specification's constraints.
\end{lstlisting}

\paragraph{Agent 9: \texttt{AgentSyntaxPDDL}.}
\begin{lstlisting}[style=prompt]
You are an agent that evaluates each step of a plan. Analyze the provided plan against the human specification and identify every inconsistency with the PDDL syntax expected by the *{target_solver}* planner. Confirm that availability and preference constraints remain encoded correctly after your edits and that the :requirements list only contains features supported by Fast Downward (:typing, :negative-preconditions, and :action-costs only when (increase ...) effects remain). Remove placeholder tokens such as '...' or 'None', and ensure action schemas never rely on ':cost' headers---instead express costs with (increase ...) effects.
\end{lstlisting}

\paragraph{Agent 10: \texttt{AgentAsynchronicity}.}
\begin{lstlisting}[style=prompt]
You are an agent that optimizes plans for asynchronicity. Analyze the provided plan against the human specification and introduce a timestamp variable that marks when each agent performs an action. Use it to schedule independent actions concurrently while keeping compatibility with the *{target_solver}* planner. Do not violate any availability or preference constraint while reshaping the schedule, and ensure you do not introduce unsupported Fast Downward features (e.g., :fluents, axioms, conditional effects). You may think through the problem in as many steps as needed.
\end{lstlisting}

\paragraph{Agent 11: \texttt{NoOpAgent}.}
Termination signal, not LLM-prompted. Fires when the validator
has accepted the current plan; returns
\texttt{[Info] NoOpAgent called. No more refinements will be performed.}
and the framework exits the refinement loop.

\paragraph{Agent 12: \texttt{AgentNaturalLanguage}.}
\begin{lstlisting}[style=prompt]
You are an agent that translates PDDL artifacts into natural language instructions. Turn the JSON specification, PDDL problem, PDDL domain, and PDDL plan into a clear sequence of human-readable actions. Follow the plan provided within the <plan></plan> tags closely. You may think through the conversion in as many steps as needed.
\end{lstlisting}

\subsection{Deterministic plan-repair agents (IDs 13--17)}
\label{app:agents-armony}

These five agents are non-LLM routines adapted from
\citet{armony2025plan}; they execute in milliseconds and contribute
no LLM cost.

\paragraph{Agent 13: \texttt{AgentVariableSwapper}.}
Fixes wrong variable bindings in PDDL plans by beam-search over
permutations of action arguments. Identifies actions with
potentially swapped arguments, enumerates candidate permutations
under a beam width, and runs each through the validator until one
succeeds.

\paragraph{Agent 14: \texttt{AgentActionReorderer}.}
Reorders actions in a plan using circular shifts and
dependency-aware swaps. Analyses precondition/effect dependencies,
detects out-of-order action pairs, and applies the minimal swap
that produces a valid ordering.

\paragraph{Agent 15: \texttt{AgentRedundancyStripper}.}
Removes redundant, duplicate, or oscillating actions. Identifies
actions that undo previous effects, duplicate consecutive actions,
and actions that do not contribute to the goal. Returns a leaner
plan.

\paragraph{Agent 16: \texttt{AgentPlanTruncator}.}
Simulates the plan from the initial state, checks goal satisfaction
at every prefix, and truncates the plan at the earliest
goal-satisfying state. Useful when the LLM produced extra actions
after the goal was already reached.

\paragraph{Agent 17: \texttt{AgentSubplanFiller}.}
Fills gaps in partial plans by invoking Fast Downward on the
sub-problem between two valid plan segments. Falls back to a full
Fast Downward call if no valid segment exists. The only
deterministic agent that itself invokes the planner.

\subsection{PDDL-aware LLM agents (IDs 18--20)}
\label{app:agents-nl2plan}

These three agents are LLM-prompted and adapted from
\citet{gestrin2024nl2plan}. Each addresses a class of PDDL-modelling
error that the baseline 13 do not cover.

\paragraph{Agent 18: \texttt{AgentTypeHierarchyFixer}.}
\begin{lstlisting}[style=prompt]
You are a PDDL expert specializing in type hierarchy design. Your task is to evaluate and fix the type hierarchy in a PDDL domain model.

Key principles:
- Each subtype should be a MORE SPECIFIC version of its parent type
- Subtypes are NOT physically contained within parent types
- Everything is ultimately a subtype of 'object'
- Types should only appear once in the hierarchy
- New meta-types can be created for organizational purposes

Use this checklist for each type:
1. Does this type fulfill the description of its parent type?
2. Are there types that should be subtypes of this type but aren't?
3. Are any types defined multiple times?

If issues are found, restructure the hierarchy and update the domain accordingly. Ensure Fast Downward compatibility with only :typing, :strips, :negative-preconditions.
\end{lstlisting}

\paragraph{Agent 19: \texttt{AgentPredicateGeneralizer}.}
\begin{lstlisting}[style=prompt]
You are a PDDL expert focused on predicate design and type generalization.

Your task is to analyze predicates in the domain and generalize their parameter types where appropriate. This helps:
- Actions work with subtypes automatically
- Reduces redundant predicate definitions
- Improves domain flexibility

When generalizing:
- Use parent types instead of specific subtypes where semantically correct
- Ensure the predicate's meaning is preserved
- Update any affected action preconditions and effects
- Don't over-generalize (e.g., don't use 'object' for everything)

Maintain Fast Downward compatibility with :typing, :strips, :negative-preconditions.
\end{lstlisting}

\paragraph{Agent 20: \texttt{AgentInitialStateSuggester}.}
\begin{lstlisting}[style=prompt]
You are a PDDL expert specialized in initial state completeness analysis.

Your task is to identify and add missing facts in the :init section of the problem that prevent the planner from finding a solution. Use auxiliary domain analysis:

1. For each action, identify what preconditions must be true
2. Trace back what initial facts would enable those preconditions
3. Check if those facts exist in :init
4. Add any missing but logically necessary facts

Important principles:
- Only add facts that MUST be true based on the specification
- Don't add facts that contradict the goal
- Don't add redundant facts
- Ensure type consistency with objects
- Consider symmetric predicates (e.g., if connected(a,b), maybe connected(b,a))

Focus on common missing patterns:
- Location/position initialization for movable objects
- Connectivity/adjacency facts
- Resource availability flags
- Object state initialization (e.g., arm-empty)

Maintain Fast Downward compatibility.
\end{lstlisting}

\section{Hyperparameters}
\label{app:hyperparameters}

We list the hyperparameters used in the headline runs reported in
Section~\ref{sec:results-main}. The same configuration is applied
to all three base models (Llama-3-8B-Instruct,
Qwen-2.5-7B-Instruct, Gemma-2-9B-it) unless otherwise noted.

\textbf{Base model and adapter configuration.}
\begin{itemize}
    \item Base precision: 4-bit NF4 quantisation.
    \item Adapter and head precision: BF16.
    \item LoRA rank $r$: 16. LoRA alpha $\alpha$: 32. LoRA dropout:
    0.05.
    \item Target modules: \texttt{q\_proj}, \texttt{k\_proj},
    \texttt{v\_proj}, \texttt{o\_proj}, \texttt{gate\_proj},
    \texttt{up\_proj}, \texttt{down\_proj}.
    \item LM head: trained (not frozen).
\end{itemize}

\textbf{Optimisation.}
\begin{itemize}
    \item Optimiser: AdamW with $\beta_1 = 0.9$, $\beta_2 = 0.999$,
    $\epsilon = 10^{-8}$.
    \item Learning rate: $2 \times 10^{-5}$.
    \item Schedule: linear with 100 warmup steps.
    \item Weight decay: $0$ on LoRA adapters.
    \item Gradient clipping: $1.0$.
    \item Per-device batch size: 4; gradient accumulation: 4
    (effective batch 16).
    \item Epochs: 3. Max sequence length: 2048 tokens.
\end{itemize}

\textbf{Data.}
\begin{itemize}
    \item Training domains: 11 PDDL domains (Section~\ref{sec:setup-benchmarks}).
    \item Seed problems: $\approx 200$.
    \item Verifier-accepted augmented trajectories:
    $\approx 5{,}000$.
    \item After LLM-as-judge soft filter: $\approx 2{,}000$.
    \item Final supervised set after exclusion of
    $\mathrm{is\_hardcoded} = \mathrm{True}$ steps:
    $\approx 12$--$15$k (state, agent) pairs.
\end{itemize}

\textbf{Inference.}
\begin{itemize}
    \item Decoding: greedy (temperature $0$, argmax over the agent
    ID token).
    \item Max refinement iterations $T_{\max}$: 10.
    \item Inference hardware: single 24\,GB consumer GPU.
    \item Latency per orchestration decision: $< 10$\,ms.
\end{itemize}

\section{State encoder details}
\label{app:state-encoder}

The state encoder renders $s_t$ into a structured prompt that
respects the active base model's chat template. We use eight
sections, each with a per-section token cap. If the assembled
prompt exceeds the 2048-token context window, left-truncation at
the section level is applied, preserving the most recent content
within each section and inserting an ellipsis marker. The
chat template (Llama-3, Qwen-2.5, or Gemma-2) is applied to the
assembled prompt before tokenisation.

\textbf{Sections, in order.}
\begin{enumerate}
    \item \emph{Task description}: the natural-language
    specification $\sigma$. Cap: 800 tokens.
    \item \emph{Current PDDL domain} $\mathcal{D}_t$ or
    ``(not yet generated)'' on cold start. Cap: 400 tokens.
    \item \emph{Current PDDL problem} $\mathcal{P}_t$. Cap: 300
    tokens.
    \item \emph{Current plan} $\rho_t$, or ``No valid plan found
    yet.'' Cap: 200 tokens.
    \item \emph{Validator errors} $e_t$ as returned by VAL. Cap:
    200 tokens.
    \item \emph{Planner logs} (most recent Fast Downward output
    fragment). Cap: 80 tokens.
    \item \emph{Agent history} $h_t$: the names of the last 10
    agents applied. Cap: 30 tokens.
    \item \emph{Available agents}: the 21-agent list with
    one-line descriptions. Cap: 200 tokens.
\end{enumerate}

The prompt terminates with the literal string
\texttt{Selected Agent:}, which is the position at which the
model emits its single agent-ID token. The loss is masked over
all prompt tokens (label $= -100$) and computed only on the
agent-ID token, so the gradient does not propagate through the
encoded state.

\section{Model family sweep}
\label{app:model-sweep}

We fine-tune three open-weights base models with the identical
recipe of Section~\ref{sec:method-sft}: the supervised set, the
state encoder, the QLoRA configuration, and the optimiser
hyperparameters are held constant across Llama-3-8B-Instruct,
Qwen-2.5-7B-Instruct, and Gemma-2-9B-it.

Differences across families fall within $\leq 2$ percentage
points on PlanBench and $\leq 3$ percentage points on the other
benchmark families, indicating that the supervised signal is not
specific to a particular base-model architecture or tokeniser.
The headline numbers reported in Section~\ref{sec:results-main}
use Llama-3-8B as the trained orchestrator.


\section{Example trajectories}
\label{app:examples}

We present one illustrative refinement trajectory per benchmark
family. Each lists the agents selected at each step and whether
the selection was made by a Layer-1 rule or by the learned
Layer-2 policy.

\subsection{PlanBench (Blocksworld, easy)}

\textbf{Specification.} Two stacking goals over three blocks.

\textbf{Trajectory.}
\begin{enumerate}
    \item $t = 0$: \texttt{AgentEmergency} (rule, empty
    domain). Initial PDDL drafted.
    \item $t = 1$: \texttt{AgentSyntaxPDDL} (rule, syntax
    error). Validator complaint about \texttt{:requirements}
    resolved.
    \item $t = 2$: \texttt{AgentHallucinations} (learned).
    Hallucinated predicate removed.
    \item $t = 3$: \texttt{NoOpAgent} (rule, $V$ accepts).
\end{enumerate}
Outcome: validator accepts after three refinement steps; one
learned-policy call.

\subsection{Natural Plan (Meeting Planning)}

\textbf{Specification.} Schedule a meeting between four
attendees across three time zones with stated busy windows and
preferences.

\textbf{Trajectory.}
\begin{enumerate}
    \item $t = 0$: \texttt{AgentEmergency} (rule).
    \item $t = 1$: \texttt{AgentDeepThinkConstraints} (learned).
    Busy-window encoding tightened.
    \item $t = 2$: \texttt{AgentTemporalConsistency} (learned).
    Ordering invariants added.
    \item $t = 3$: \texttt{NoOpAgent} (rule).
\end{enumerate}
Outcome: three refinement steps; two learned-policy calls.

\subsection{Classical planning (Hanoi, medium, five disks)}

\textbf{Specification.} Move five disks from peg A to peg C
under standard Tower-of-Hanoi rules.

\textbf{Trajectory.}
\begin{enumerate}
    \item $t = 0$: \texttt{AgentEmergency} (rule).
    \item $t = 1$: \texttt{AgentFastDownwardsAdapter} (learned).
    Domain rewritten to remove an unsupported requirement.
    \item $t = 2$: \texttt{AgentSubplanFiller} (learned). Fast
    Downward used to fill a partial plan.
    \item $t = 3$: \texttt{NoOpAgent} (rule).
\end{enumerate}
Outcome: three refinement steps; two learned-policy calls.

\section{Cost calculations}
\label{app:costs}

\textbf{Methodology.} For prompted-orchestrator baselines we
count input and output tokens for every orchestrator forward pass
plus the tokens consumed by each agent LLM call, multiply by the
respective provider's per-token rate, and sum over all problems
in a benchmark family. For the trained orchestrator we treat
orchestration cost as the marginal cost of running the local
model forward pass; the dominant reduction comes from avoiding
the prompted-orchestrator's full-state prompt at every refinement
step. Deterministic plan-repair agents (IDs~13--17) contribute
zero LLM-cost in both pipelines because they do not call any LLM.

Per-task averages used in Section~\ref{sec:results-main} are
\$0.18--\$0.22 for the prompted GPT-5-mini orchestrator (the
higher end on Natural Plan, where prompts are longer), \$0.06--\$0.08
for the prompted Gemini-3-Flash orchestrator under the same
distribution, and \$0.004 for HALO running locally. HALO's
per-task cost is dominated by the amortised local-GPU operating
cost rather than per-token API charges, which is what underwrites
the roughly 45$\times$ reduction against GPT-5-mini (and
15$\times$ against Gemini-3-Flash).



\end{document}